\documentclass[letterpaper]{article} 
\usepackage{aaai23}  
\usepackage{times}  
\usepackage{helvet}  
\usepackage{courier}  
\usepackage[hyphens]{url}  
\usepackage{graphicx} 
\usepackage{natbib}  
\usepackage{caption} 
\usepackage{algorithm}
\usepackage{algorithmic}
\usepackage[title,toc,titletoc,page]{appendix}

\usepackage{newfloat}
\usepackage{listings}
\DeclareCaptionStyle{ruled}{labelfont=normalfont,labelsep=colon,strut=off} 
\floatstyle{ruled}
\newfloat{listing}{tb}{lst}{}
\floatname{listing}{Listing}
\usepackage{algorithm}
\usepackage{algorithmic}
\usepackage{bm}
\usepackage{amsthm}
\usepackage{amsfonts}

\usepackage{multirow}

\usepackage{booktabs}
\usepackage{subcaption}

\usepackage{amsmath}
\usepackage{cleveref}

\title{Supervised Contrastive Prototype Learning: \\Augmentation Free Robust Neural Network}
\author {
    Iordanis Fostiropoulos,
    Laurent Itti
}
\affiliations{
    University of Southern California\\
    fostirop@usc.edu, itti@usc.edu
}

\usepackage{bibentry}

\begin{document}

\maketitle

\begin{abstract}

Transformations in the input space of Deep Neural Networks (DNN) lead to unintended changes in the feature space. Almost perceptually identical inputs, such as adversarial examples, can have significantly distant feature representations. On the contrary, Out-of-Distribution (OOD) samples can have highly similar feature representations to training set samples. 
Our theoretical analysis for DNNs trained with a categorical classification head suggests that the inflexible logit space restricted by the classification problem size is one of the root causes for the lack of \textit{robustness}. Our second observation is that DNNs over-fit to the training augmentation technique and do not learn \textit{nuance invariant} representations. Inspired by the recent success of prototypical and contrastive learning frameworks for both improving robustness and learning nuance invariant representations, we propose a training framework, \textbf{Supervised Contrastive Prototype Learning} (SCPL). We use N-pair contrastive loss with prototypes of the same and opposite classes and replace a categorical classification head with a \textbf{Prototype Classification Head} (PCH). 
Our approach is \textit{sample efficient}, does not require \textit{sample mining}, can be implemented on any existing DNN without modification to their architecture, and combined with other training augmentation techniques. We empirically evaluate the \textbf{clean} robustness of our method on out-of-distribution and adversarial samples. Our framework outperforms other state-of-the-art contrastive and prototype learning approaches in \textit{robustness}.

\end{abstract}

\begin{figure}[htp]
\centering
\begin{subfigure}[b]{1\linewidth}
\centering
\includegraphics[width=.5\linewidth]{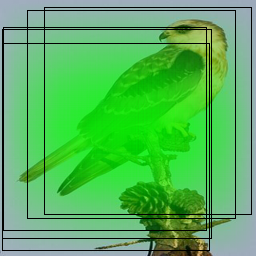}
\caption{Random-Crop Probability Heatmap}
\label{fig:aug_map}
\end{subfigure}\\
\begin{subfigure}[b]{0.49\linewidth}
\includegraphics[width=1\linewidth]{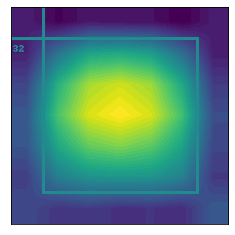}
\caption{ImageNet Activation Map}
\label{fig:imgnets_grad}
\end{subfigure}
\begin{subfigure}[b]{0.49\linewidth}
\includegraphics[width=1\linewidth]{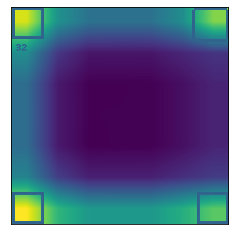}
\caption{Noise Activation Map}
\label{fig:rand_grad}
\end{subfigure}
\centering
\caption{ \textbf{Top}: (\subref{fig:aug_map}) The heatmap of the random-crop where the color intensity signifies high probability sample regions. Five exemplar crop bounding boxes are also shown. \textbf{Bottom}: Artifact of over-fitting the training augmentation technique for a VGG11 model pre-trained on ImageNet. The pixel feature importance is obtained using GradCam \cite{selvaraju2017grad}. 
The mean feature importance of the validation set is calculated for (\subref{fig:imgnets_grad}) in-distribution ImageNet and (\subref{fig:rand_grad}) out-of-distribution Gaussian random noise.
During inference, the model uses highly probable features for classification of the in-distribution dataset, while it uses most difficult features for out-of-distribution predictions. Detailed explanation in \cref{sec:intro} and Appendix sec. B.
}\label{figure:activation_maps}
\end{figure}

\section{Introduction}
\label{sec:intro}

DNNs are known for their poor performance in generalizing to unseen inputs or simple image transformations, such as rotation \cite{nguyen2015deep,guo2017calibration}. 
Two inputs with insignificant $l_p$ distance in the input space can produce significantly distant feature representations that lead to misclassification. On the contrary, input that is perceptually different from the training samples, such as random noise, can result in feature representations nearly identical to the training samples. In this work, we use the term \textbf{robustness} to refer to the property of DNN to learn an equidistant projection from the input space to the feature space. We identify two main failure cases in DNN; lack of robustness against \textit{perturbations}, and against \textit{nuances}. 

\begin{figure*}[htp]
\centering
\includegraphics[width=.9\linewidth]{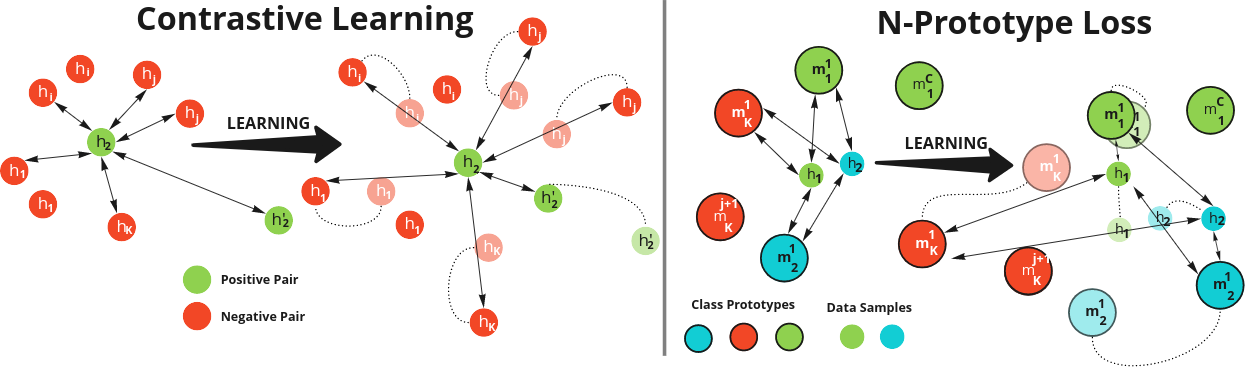}
\centering
\caption{NT-Xent Loss \cite{Npair} (left) updates DNN weights to encourage feature vectors $h_i$ of class $i$ to be closer to their intra-class samples based on cosine similarity. Ineffective sample mining can lead to inter-class entanglement such as for \textit{hard negatives} $h_1$ and $h_2$. N-Prototype Loss (right) applies the loss between the closest prototype vector $\bm{m}_i^j$ for class $i$. $h_1$ and $h_2$ do not explicitly interact. The optimization between class representative prototypes and samples implicitly leads to inter-class separation between samples. }
\label{figure:learning}
\end{figure*}

Robustness against perturbations can be evaluated with an input crafted by an adversarial attack, where an insignificant perturbation on the input space can cause a significant perturbation on the feature space \cite{szegedy2013intriguing}. Similarly, robustness against \textit{nuances} can be observed with the weakness of DNN where they assign nearly identical feature representations to input that is significantly distant in the $l_p$ input space, such as between random noise and training samples.

\begin{figure}[htp]
\centering
\includegraphics[width=1\linewidth]{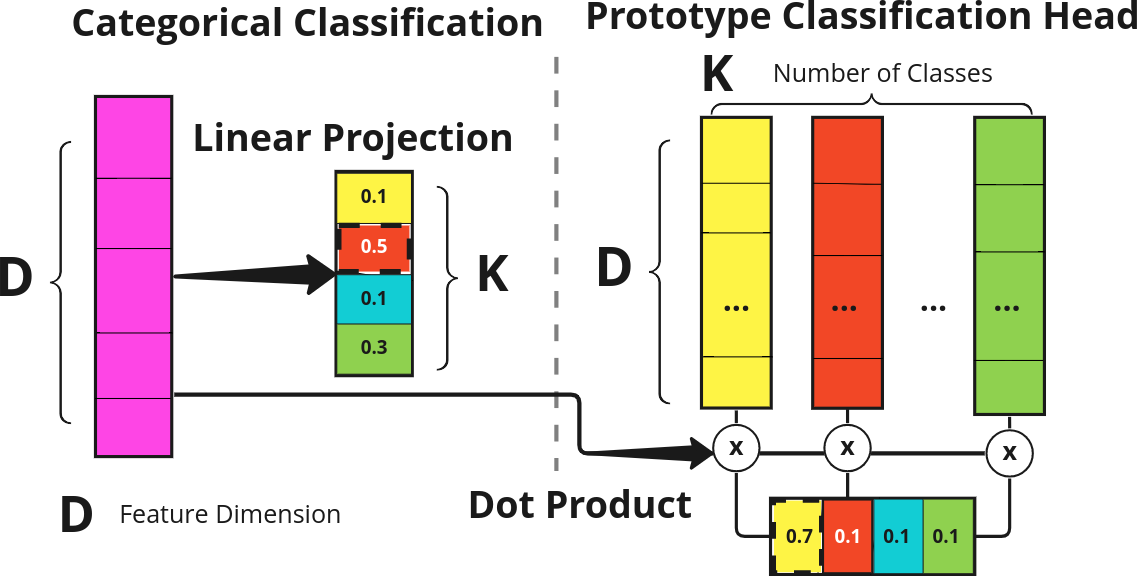}
\centering
\caption{Categorical Classification Head (left) projects the penultimate feature vector to a K-Dimensional vector that is the same dimensionality as the number of classes. Our method, Prototype Classification Head (right) performs prototype matching where the feature dimensionality is flexible. Our analysis shows that  increasing the feature dimension $D$ provides stronger robustness guarantees. }
\label{figure:cat_learning}
\end{figure}

Previous work improves the robustness of DNNs with training augmentation techniques specific to the problem they solve. For example, some work \cite{papernot2016distillation,madry2017towards,papernot2017practical} improves adversarial robustness by including adversarial-generated samples in the training set. OOD detection methods use samples from the OOD set \cite{lee2017training,liang2017enhancing} or an auxiliary dataset \cite{hendrycks2018deep} during training.

Previous work in contrastive learning \cite{simclr} identifies training augmentation techniques as one of the primary contributing factors to learning robust feature representations invariant to nuances in the input. In this work, we also view that training augmentation techniques are necessary for learning representations invariant to noise. However, we identify two issues with current approaches in training augmentation. 

First, DNNs are prone to over-fitting.
We find that,  for multiple architectures, the majority of OOD samples are assigned to the most difficult class, the one with the least salient features. Our observation is at odds with the goal of a classifier to learn salient features for every class. For example, for a pretrained ResNet classifier, 99\% of uniform random noise is assigned to class 104 with high probability (Appendix fig. S2 c). This suggests a failure of learning, whereby all kinds of unknown images are assigned to the most uncertain class (here, 104). Additionally, DNNs over-fit the augmentation technique used to train the classifier. 
Previous work \cite{he2016deep,simonyan2014very} resize training images to 256 pixels on the shortest side and then randomly crop them to (224x224) pixels when training on ImageNet. \cref{figure:activation_maps} visualizes artifacts of over-fitting to the augmentation technique used during training. The model is biased to the least salient features when predicting out-of-distribution images.

Secondly, data augmentation techniques are expensive, especially for OOD methods that require samples from an out-of-distribution set
or an auxiliary dataset. Such methods can be limited in practical settings where additional data is scarce. Similarly, generating adversarial examples for every training iteration is expensive and the current state-of-the-art \cite{wong2020fast} focuses on improving computational performance. Current training augmentation solutions are a quick fix to a systematic problem, where the issue of robustness is an issue of generalization \cite{xu2012robustness}.

Contrastive Learning \cite{hadsell2006dimensionality} approaches do not directly solve for a task but optimize a similarity metric such as the euclidean distance or cosine similarity between feature vectors. Previous contrastive learning work has been successful on downstream tasks such as classification, zero-shot learning, and more \cite{wang2018cosface,xu2020attribute,wang2017deep,chen2017beyond}. Additionally, the metric space learned by such models can be robust to both adversarial examples and OOD. 
The focus of state-of-the-art contrastive learning approaches \cite{chen2020simple,khosla2020supervised} has been to find effective ways to sample pairs that are of the same class (``positive pair'') and one of the most-similar other classes (``negative pair'') that produce meaningful gradients. The \textit{sample mining} process is computationally expensive and the downstream task performance is sensitive to both the sampling method and the sample size \cite{kaya2019deep}

Prototypical Learning (PL) uses a set of vectors (\textit{prototypes}) to represent a class and is an extension of k-means, where each of the K vectors is a prototype. PL has been used in combination with different learning paradigms such as pattern recognition \cite{liu2001evaluation} and more recently has been used both for zero-shot and few-shot learning due to the generalization and sample efficiency \cite{snell2017prototypical}. Frameworks have been proposed that implement PL in both supervised and unsupervised settings  \cite{DCE,li2020prototypical}.

 Our goal is to design a model that does not depend on any training augmentation technique or auxiliary datasets to achieve high robustness. Motivated by the recent success of Contrastive Learning and Prototypical Learning approaches, in this work we propose the \textbf{Prototype Classification Head} (PCH). Our proposed method combines a classification task with a \textbf{purely} contrastive, sample efficient loss, and can be trained end-to-end with a DNN backbone. PCH is used at the penultimate feature representation of a DNN and replaces the classification head. We use a set of prototype vectors to represent a cluster of feature vectors of the same class. We further improve on previous prototype learning approaches and use N-Prototype Loss which increases inter-class separability and intra-class compactness among training samples. Our approach creates more robust models without additional tricks. Our contribution in detail:

\begin{itemize}
    \item To the best of our knowledge, this is the first work to combine prototype learning with a purely contrastive loss between prototypes and samples, \textit{N-Prototype Loss}
    \item Compared to traditional prototype and contrastive learning methods, our framework uses prototypes to represent a class and does not require intra-batch sample mining.
    \item Our proposed framework is flexible and can be added to any existing DNN architectures without any further modifications.
    \item We provide a theoretical justification for the improved robustness, and conduct quantitative experiments that outperform other state-of-art methods
\end{itemize}

\section{Related Work}

We identify Prototype Learning approaches and Contrastive Learning approaches as two closely related categories of work that improve the robustness of DNN classifiers.

Triplet loss \cite{hermans2017defense} uses one negative and positive pair for each sample to learn a metric and can be sensitive to noise which leads to over-fitting. N-pair\cite{Npair} loss decreases the similarity between multiple negative pairs and increases the similarity between a positive pair. Current contrastive learning approaches suffer from poor quality sample pairs that generate uninformative sub-gradients which lead to poor model convergence. Mining for informative pairs such as \textit{hard negatives} can be computationally expensive.  Several approaches have been proposed to improve on the \textit{sample mining} problem. \cite{duan2018deep} train a separate negative sample generator jointly with a DNN that augments ``easy'' negative samples with ``hard'' negative samples. Lifted Structured Loss (LSL) \cite{oh2016deep} uses multiple positive and negative pairs in the importance sampling method to improve \textit{sample mining} of \textit{hard negative} pairs. However, sample mining techniques are computationally expensive. In contrast, we use learned prototypes for positive and negative pairs and thus avoid the \textit{sample mining} problem.

Contrastive learning losses are applied in a supervised setting. \cite{TLA} use Triplet loss in an adversarial training setting as an auxilary training objective to the cross-entropy loss. 
Several contrastive learning framework such as SimCLR \cite{simclr}, MoCo \cite{he2020momentum} and BYOL \cite{grill2020bootstrap} use a generalized version of N-pair loss (InfoNCE) on the feature representation of an augmented sample to create a positive pair and treat all other samples in a batch as negative. 
Work by \cite{li2020prototypical} add an additional loss term to the NCE loss that compares the original sample with prototypes of the same cluster and \textit{r} negative prototypes. Both the positive and negative prototypes are determined using an expectation maximization clustering framework at every training iteration. In contrast, we assign prototypes to each corresponding class in a supervised setting. Additionally, we use a single training objective between each sample and multiple prototypes, and have a flexible penultimate feature dimension which we identify to be the root cause of the improved robustness. Lastly, our method is orthogonal to other unsupervised contrastive learning approaches and any training augmentation techniques or specific backbone architectures \cite{simclr,he2020momentum,grill2020bootstrap} can be extended to our framework.


\cite{mustafa2019adversarial} apply Prototype Conformity Loss (PCL) as the cross-entropy of the cosine similarity between the feature representation of a sample and the true class prototype vector. For every intermediate representation, an auxiliary network is trained to enforce the intra-class compactness and inter-class separation using the learned prototypes. In contrast to our method, it requires modifying the existing network architecture and can not be directly applied to any existing architecture. Additional work \cite{mettes2019hyperspherical,chen2021adversarial} define fixed prototypes based on an inductive bias for the latent space geometry and maximize the cosine similarity between a sample and the positive class prototype. In contrast to our method, the loss function is applied between multiple sample feature representations and with multiple prototypes from the positive and negative class. Lastly, in our framework the prototypes are learned end-to-end with the DNN. Maintaining fixed prototypes is orthogonal to our method and can be considered as a special case of our framework.

The work that is most similar to ours is \cite{yang2018robust} that use Distance-based Cross-Entropy (DCE) loss and \cite{liu2020convolutional} that extend on DCE with an additional loss term Prototype Encoding Cost (PEC). DCE is applied between the closest positive class prototype and a sample, while PEC is the $l_2$ norm between the positive class prototype and the penultimate feature representation. DCE and PEC are optimized jointly for both the network parameters and learned prototypes. Similarly, our framework is applied to the penultimate feature representation. Contrary to DCE and PEC, at every step our method increases the distance between a sample and multiple ($N$) opposite class prototypes and reduces the distance to the closest positive prototype.

There are additional work \cite{yang2020convolutional,shu2020p} that use prototype learning for a specific problem but can be seen as derivation of more general frameworks. In this work and for reasons of clarity we provide a generalized definition of the state-of-the-art approaches in \cref{sec:preliminary}.
Extensive experiments show that our method outperforms other state-of-the-art \cite{TLA,simclr,DCE} methods in \textit{clean robustness} and does not require training augmentation techniques to achieve high robustness. Finally, when compared to other metric learning approaches, our approach does not require sample mining.

\section{Preliminary}
\label{sec:preliminary}
In this work, we consider the issue of robustness from two angles. Robustness against data samples crafted by an adversary and robustness against samples that are out-of-distribution compared to the training data set.\\

\textbf{Out-of-Distribution Robustness} (OOD) evaluates a classifier \(f_\theta(x)=h_x\) trained on a dataset  \(\mathcal{D}^\emph{in}\) with an auxiliary function $g(h_x)$ on a test dataset \(\mathcal{D}^\emph{out}\) with \(\mathcal{D}^\emph{out}\cap \mathcal{D}^\emph{in}=\emptyset \) such that
\begin{equation}
\label{eq:ood}
 g(h_x) = \left\{
\begin{array}{ll}
     1 & x \in \mathcal{D}^\emph{out} \\
     0 & x \in \mathcal{D}^\emph{in}
\end{array}
\right.
\end{equation}
In this work, we do not design a specific method to our model for OOD detection and use work by \cite{OODbaseline}. 
\\
\textbf{OOD Baseline} \cite{OODbaseline} normalize the predicted class similarity for a sample $x$ and classifier \(f_\theta(x)\) and use the maximum value of the prediction scores for all classes $K$ as the confidence score \(p(h_x)=\max{\sigma(h_x)}\)
\begin{equation}
\label{eq:ood_baseline}
 g(h_x) = \left\{
\begin{array}{ll}
     1 & p(h_x)\leq \alpha \\
     0 & p(h_x) > \alpha
\end{array}
\right.
\end{equation}\\
where a confidence threshold $\alpha$ is used to determine if $x\in D^\emph{out}$. The robustness of a classifier is then determined by the performance of $g(f_\theta(x))$ in correctly classifying in and out-of-distribution samples for a fixed $\alpha$.

\textbf{Adversarial Robustness} measures the accuracy of a DNN with \(f_{\theta}\) in correctly predicting a clean sample $x$, after a perturbation $\delta$ has been applied by an adversary, with perturbation budget $\varepsilon$ such that

\begin{equation}
\label{eq:adv_robustness}
    \underset{\delta}{\mathrm{argmax}} \enspace \mathcal{L}(x + \delta, y), \enspace s.t., \parallel\delta\parallel_{\textit{p}} < \varepsilon,
\end{equation}
where \(\mathcal{L}(\cdotp)\) denotes the loss function used to train \(f_{\theta}\) and \(\parallel\delta\parallel_{\textit{p}}\) is the $\textit{p}$-norm of $\delta$ with \(p \geq 1\). As such, the adversarial example lies within an \(\ell_{p}\)-ball centered at $x$.

A robust model can correctly classify a perturbed sample that is within the perturbation budget. In this work, we consider adversaries that can manipulate a data sample in $l_\infty$ and $l_2$ space. \\
We consider the following adversarial attack settings: Fast Gradient Sign Method (\textit{FGSM}) \cite{goodfellow2014explaining},
Basic Iterative Method (\textit{BIM}) \cite{AT}
, Projected Gradient Descent (\textit{PGD}) \cite{madry2017towards} and AutoAttack \cite{croce2020reliable}.

\subsection{Prototype Distance based Cross Entropy (DCE)}\label{sec:dce} 

Prototypical frameworks  \cite{yang2018robust,liu2020convolutional} minimize the distance between prototypes $m_i$ for a task with K classes and sample feature vectors $h_x$. A similarity matrix between a given feature vector and all prototypes is normalized  to satisfy the non-negative and sum-to-one properties of probabilities such that
\begin{equation}
\label{eq:dce}
\displaystyle
p(y|x)=\frac{e^{\gamma sim(h_x, m_y)}}{\sum\limits_{i \in K} e^{\gamma sim(h_x, m_i)}}
\end{equation}
where y is the true class of sample $x$ and $\gamma$ is a hyper-parameter. Objective \cref{eq:dce} is a generalization of the traditional categorical cross-entropy, when the cosine similarity is used as a similarity function and the prototypes $m_i$ are fixed one-hot vectors. 

\subsection{Auxilary Contrastive Loss}\label{sec:atl} 
There are work \cite{TLA} that use a contrastive loss that is \textit{auxilary} to the primary categorical cross-entropy classification loss. 

\textbf{Auxilary Triplet Loss} ($ATL$) \cite{chen2017beyond} denoted by $\mathcal{L}_{trip}$ can be used for classification tasks when optimized jointly with a cross-entropy loss and is computed by \cref{eq:tla}
\begin{equation}
\label{eq:tla}
 \mathcal{L}_{CE}(f(x), y) +\lambda_1\mathcal{L}_{trip}(h_{x}, h_{y}, h_{\bar{y}})+\lambda_2\mathcal{L}_{norm}
\end{equation}
with  \(\mathcal{L}_{norm} = ||h_{x}||_2 +||h_p||_2 + ||h_n||_2\) and where $h_{x}$ is a feature representation of $x$, and $h_y$ and $h_{\bar{y}}$ are the feature representations of a positive and negative sample to $x$ respectively. $\lambda_1$ is a positive coefficient that provides a trade-off between $L_{CE}$ and $L_{trip}$, and $\lambda_2$ is the weight for the feature norm. The formulation of the ATL in this section is a generalization of \cite{TLA} but without the use of adversarially augmented samples.

\textbf{Auxilary N-pair Loss (ANL)}, 
\label{sec:anl} We extend ATL by replacing \(\mathcal{L}_{trip}\) with an InfoNCE Loss Objective \cite{sohn2016improved} in \cref{eq:tla} where $L_{npair}$ is calculated on a (N+1)-tuple \((h_{x}, h_y, \{{h_{\bar{y}}}\})\), and \(\{{h_{\bar{y}}}\}\) is the set of $N-1$ feature representations from negative samples to $x$. $\mathcal{L}_{npair}$ denotes the total loss calculated by \cref{eq:L_n_pair}
\begin{equation}
\label{eq:L_n_pair}
\log  [1 + \sum_{i=1}^{N-1}\exp (\textit{sim}(h_{x'}, {h_{\bar{y}}}^i) - \textit{sim}(h_{x'}, h_y))~]
\end{equation}
where $sim$ is the similarity function and all samples are mined identical to ATL. ANL is an interpretation of SimCLR, MoCo and BYOL but applied in a supervised setting and without any training augmentation techniques.

\section{Method}\label{sec:method}
\textbf{Supervised Constrastive Prototype Learning} (SCPL) uses a DNN \(f_\theta(x)=h_x\) as feature extractor, where $x$ is the raw input, $\theta$ the parameters of the model and $h_x$ the learned feature representation of $x$. We apply a \textit{Prototype Classification Head} on the hidden feature representation $h_x$ of $x$.

The \textbf{Prototype Classification Head} (PCH) is composed of multiple learnable prototype vectors \(\mathcal{M} = \{m_i^j \in R^D | i\in K;j\in C_i\}\) where \(i\) is the index from the set of classes ${K}$, $j$ is the index from the set of prototypes for class $i$ and $C_i$ is the total number of prototypes of the same class. For our experiments, we use the same number of prototypes $C$ for all classes, but this assumption can be relaxed for future works.

\subsubsection{Inference}
We classify a sample by \textit{prototype-matching}. Given input $x$ we compare the extracted representation $h_x$ with all prototypes and assign it to the class with index $i$ of the \textit{closest prototype} $\bm{m}_i$ based on a metric $\textit{sim}$ such that

\begin{equation}
    \bm{m}_i(x) = -\min_{m_i^j \in \mathcal{M}} \textit{sim}(h_x,m_i^j)
\end{equation}

\subsubsection{Learning}
The trainable parameters in our framework are the parameters  $\theta$ for the DNN and the prototypes for each class \(\mathcal{M}\) which are optimized jointly in an end-to-end manner with a single loss function such that
\begin{equation}
    \mathcal{L}_{SCPL} = \mathcal{L}_{pl} + \lambda\mathcal{L}_{norm}
\end{equation}
where \(\mathcal{L}_{pl}\) is the \textit{N-Prototype Loss}, \(\mathcal{L}_{norm}\) the \textit{Prototype Norm} and $\lambda$ is a hyper-parameter that controls the weight of the prototype norm. The proposed framework can be extended to other metric learning loss functions such as Contrastive, Triplet, and N-pair. Instead of calculating a metric between samples, we calculate the same metric between prototype vectors and the feature representation of samples. For our experiments, we use N-Prototype Loss.\\
\textbf{N-Prototype Loss} is computed between the feature representation $h_x$ of the training sample $x$ with label $y$ and the closest prototype vector from the positive class $\bm{m}_y$ over the set of the closest prototypes from all negative classes \(\bar{\bm{\mathcal{M}}}=\{\bar{\bm{m}}_i | i\in K; i\neq y\}\) where $\mathcal{L}_{pl}$ denotes the total loss calculated by \cref{eq:L_pl}
\begin{equation}
\label{eq:L_pl} \log [1 + \sum_{\bar{\bm{m}_{i}} \in \bar{\bm{\mathcal{M}}}}\exp (\textit{sim}(h_x, \bar{\bm{m}_{i}})-\textit{sim}(h_x,\bm{m}_y)]
\end{equation}
\textbf{Prototype Norm} further improves the compactness of learned prototypes by adding a regularization term to the objective loss function such that
\begin{equation}
\label{eq:L_norm}
    \mathcal{L}_{norm} = ||h_x - \bm{m}_y||_2
\end{equation}
The prototype norm serves the same purpose as an $l_2$-norm regularization objective to reduce the complexity of the solution space for a $\textit{sim}$ metric. Thus, $\mathcal{L}_{norm}$ improves intra-class compactness and inter-class separability by pulling a feature vector closer to the true class prototype. A difference between \cref{eq:L_n_pair} and \cref{eq:L_pl} is in the use of prototypes as opposed to negative and positive sample pairs and is highlighted in \cref{figure:learning}.

\subsubsection{Provably Robust Framework} A categorical classifier learns a decision boundary on a feature space that has the same dimensionality as the size of the classification task. For example, a four-way classification task would produce a four-dimensional feature space, for which each dimension is a feature that represents the probability that a sample belongs to each respective class. We identify the cause of the robustness-accuracy trade-off \cite{tsipras2018robustness} to be the direct relationship between the feature space dimensionality restricted by the number of classes (``problem size''). An illustration on the difference between categorical classification head and PCH is highlighted in \cref{figure:cat_learning}.

PCH has a flexible $D$ independent of the problem size dimensionality. This allows us to show that the improved performance is a consequence of the flexible dimensionality of the learned feature representation for our method and provide a detailed proof in the Appendix sec. A.

Our experiments corroborate the theoretical implications. We can increase the robustness of the model by increasing $D$ and 
models robust to Gaussian noise are also robust to adversarial attacks. For reasons of brevity we provide additional analysis and experiments that validate our theoretical findings in the Appendix sec. B.
\section{Experiments}
\label{sec:exprm}
\begin{table}
\caption{Adversarial Robustness on MNIST, CIFAR10, and SVHN datasets. Attacks for the MNIST dataset are 20 steps with $\varepsilon=0.3$, and 7 steps for CIFAR and SVHN datasets with $\varepsilon=\frac{8}{255}$.}
\label{table:adv_robustness}
\begin{center}
\scalebox{0.7}{

\begin{tabular}{@{}llcccccc@{}}
\toprule

& & \textbf{Clean}&\textbf{FGSM}&\textbf{PGD}& \textbf{BIM}& $\textbf{PGD}_{RS}$& $\textbf{PGD}_{L2}$\\
\midrule
\multirow{5}{*}{\textbf{MNIST}} & LeNet & 99.39   & 11.9    &  1.66     & 76.98     & 26.31     & 99.01   \\
& DCE & 99.02   & 11.75   &  0.72     & 73.12     & 25.70      & 98.43   \\
& ATL & 99.32   & 42.13   &  9.75     & 75.54     & 33.69     & 98.85   \\
& ANL & 99.42   & 48.13   &  4.42     & 76.99     & 49.09     & 99.06   \\
\cmidrule(l){2-8}
& \textbf{Ours} & \textbf{99.45}   & \textbf{65.65}   &  \textbf{27.65}    & \textbf{90.02}     & \textbf{79.03}     & \textbf{99.23}   \\
\midrule
\multirow{5}{*}{\textbf{CIFAR}} & ResNet & 82.76   & 10.83   &  0.10      &  0.14     &  0.48     & 56.93   \\
& DCE   & 82.28   & 26.37   &  6.37     &  8.81     & 10.07     & 55.83   \\
& ATL    & 83.45   & 16.4    &  0.39     &  0.72     &  1.96     & 57.94   \\
& ANL  & 83.89   & 13.85   &  3.59     &  4.23     &  6.26     & 56.24   \\
\cmidrule(l){2-8}
& \textbf{Ours}     & \textbf{88.980}  & \textbf{56.8}    & \textbf{31.56}     & \textbf{31.23}     & \textbf{44.86}     & \textbf{71.47}   \\
\midrule
\multirow{5}{*}{\textbf{SVHN}}  & ResNet   & 92.13 & 16.95 &  0.14 &  0.14 &  0.71 & 63.56 \\
& DCE & 92.17 & 50.99 &  5.64  &  8.23  & 17.39   & 72.25 \\
& ATL & 92.78 & 18.85 &  1.10  &  1.41   &  3.41  & 72.08 \\
& ANL & 92.85 & 34.14 & 23.18   & 23.60   & 23.82   & 70.34 \\
\cmidrule(l){2-8}
& \textbf{Ours}     & \textbf{94.9} & \textbf{81.06} & \textbf{56.10}    & \textbf{58.02}   & \textbf{71.90}     & \textbf{88.08} \\ \bottomrule
\end{tabular}
}
\end{center}
\end{table}
\subsection{Datasets and Models}
 We extensively evaluate the proposed method on three datasets, MNIST \cite{lecun2010mnist}, CIFAR10 \cite{Krizhevsky09learningmultiple}, and Street-View House Numbers(SVHN) \cite{Netzer2011}. We use Ray \cite{liaw2018tune} optimization framework on a GPU cluster. 
 We provide larger figures and additional experiments in the supplementary material.

We use the categorical cross-entropy loss function for all baselines. As a \textit{backbone} we use ResNet-18 \cite{he2016deep} for CIFAR10 and SVHN datasets and LeNet \cite{lecun1989backpropagation} for MNIST.

For baselines, we consider metric learning and convolutional prototype learning approaches as described in \cref{sec:preliminary}.
We compare our method with Auxilary Triplet Loss ``ATL'' from \cref{sec:atl} used by \cite{TLA}. We also compare with Auxilary N-pair Loss ``ANL'' from \cref{sec:anl} where we use one negative pair per class in $L_{npair}$ which is the interpertation of SimCLR, BYOL and MoCo for a supervised-learning setting. Finally we compare with generalized convolutional prototype learning (GCPL) framework \cite{DCE, liu2020convolutional} and denote it as ``DCE''.

and use a dimension \(D\) of 64, and a \(\lambda\) of 0.1 for $L_{pl}$.

\subsection{Results}
\subsubsection{Clean Robustness}

\begin{table}[h]
\begin{center}
\caption{Out-of-distribution detection on ImageNet, LSUN datasets, and Gaussian noise inputs with SVHN as in-distribution datasets.}
\label{table:ood_svhn}
\scalebox{0.65}{
\begin{tabular}{@{}clccccc@{}}
\toprule
                                                                                   & \multicolumn{1}{c}{} & \textbf{\begin{tabular}[c]{@{}c@{}}FPR\\ (95\% TPR)\\ $\downarrow$ \end{tabular}} & \textbf{\begin{tabular}[c]{@{}c@{}}Detection\\ Error\\ $\downarrow$ \end{tabular}} & \textbf{\begin{tabular}[c]{@{}c@{}}AUROC\\ \\ $\uparrow$ \end{tabular}} & \textbf{\begin{tabular}[c]{@{}c@{}}AUPR\\ In\\ $\uparrow$\end{tabular}} & \textbf{\begin{tabular}[c]{@{}c@{}}AUPR\\ Out\\ $\uparrow$\end{tabular}} \\ \midrule
\multirow{5}{*}{\textbf{ImageNet}}                                            & ResNet              & 86.8\%                                                                & 31.4\%                                                                 & 73.8\%                                                        & 72.9\%                                                         & 68.1\%                                                          \\
                                                                                   & DCE                  & 89.2\%                                                                & 16.7\%                                                                 & 86.4\%                                                        & 90.4\%                                                         & 75.9\%                                                          \\
                                                                                   & ATL                  & 42.5\%                                                                & 11.4\%                                                                 & 93.9\%                                                        & 95.4\%                                                         & 91.5\%                                                          \\
                                                                                   & ANL                  & 62.7\%                                                                & 19.3\%                                                                 & 86.4\%                                                        & 84.0\%                                                         & 82.6\%                                                          \\ \cmidrule(l){2-7}
                                                                                   & \textbf{Ours}        & \textbf{9.9\%}                                                        & \textbf{5.3\%}                                                         & \textbf{96.3\%}                                               & \textbf{97.7\%}                                                & \textbf{91.7\%}                                                 \\ \midrule
\multirow{5}{*}{\textbf{LSUN}}                                                     & ResNet              & 68.5\%                                                                & 20.1\%                                                                 & 86.4\%                                                        & 87.2\%                                                         & 82.3\%                                                          \\
                                                                                   & DCE                  & 94.0\%                                                                & 21.5\%                                                                 & 80.6\%                                                        & 83.4\%                                                         & 70.2\%                                                          \\
                                                                                   & ATL                  & 57.3\%                                                                & 15.2\%                                                                 & 90.8\%                                                        & 92.6\%                                                         & 87.5\%                                                          \\
                                                                                   & ANL                  & 53.2\%                                                                & 18.0\%                                                                 & 87.1\%                                                        & 82.2\%                                                         & 85.0\%                                                          \\ \cmidrule(l){2-7}
                                                                                   & \textbf{Ours}        & \textbf{9.2\%}                                                        & \textbf{5.6\%}                                                         & \textbf{96.8\%}                                               & \textbf{97.8\%}                                                & \textbf{94.6\%}                                                 \\ \midrule
\multirow{5}{*}{\textbf{\begin{tabular}[c]{@{}c@{}}Gaussian\\ noise\end{tabular}}} & ResNet              & 92.5\%                                                                & 44.3\%                                                                 & 57.8\%                                                        & 49.5\%                                                         & 53.1\%                                                          \\
                                                                                   & DCE                  & 23.0\%                                                                & 7.6\%                                                                  & \textbf{96.3\%}                                               & 97.4\%                                                         & \textbf{93.7\%}                                                 \\
                                                                                   & ATL                  & 36.9\%                                                                & 10.8\%                                                                 & 94.7\%                                                        & 95.8\%                                                         & 92.7\%                                                          \\
                                                                                   & ANL                  & 97.3\%                                                                & 50.0\%                                                                 & 41.0\%                                                        & 41.7\%                                                         & 45.0\%                                                          \\ \cmidrule(l){2-7}
                                                                                   & \textbf{Ours}        & \textbf{4.8\%}                                                        & \textbf{2.8\%}                                                         & 95.5\%                                                        & \textbf{97.7\%}                                                & 86.2\%                                                          \\ \bottomrule
\end{tabular}
}
\end{center}
\end{table}
\begin{figure*}[h]
\centering
\begin{subfigure}{0.44\textwidth}
\includegraphics[width=1\textwidth]{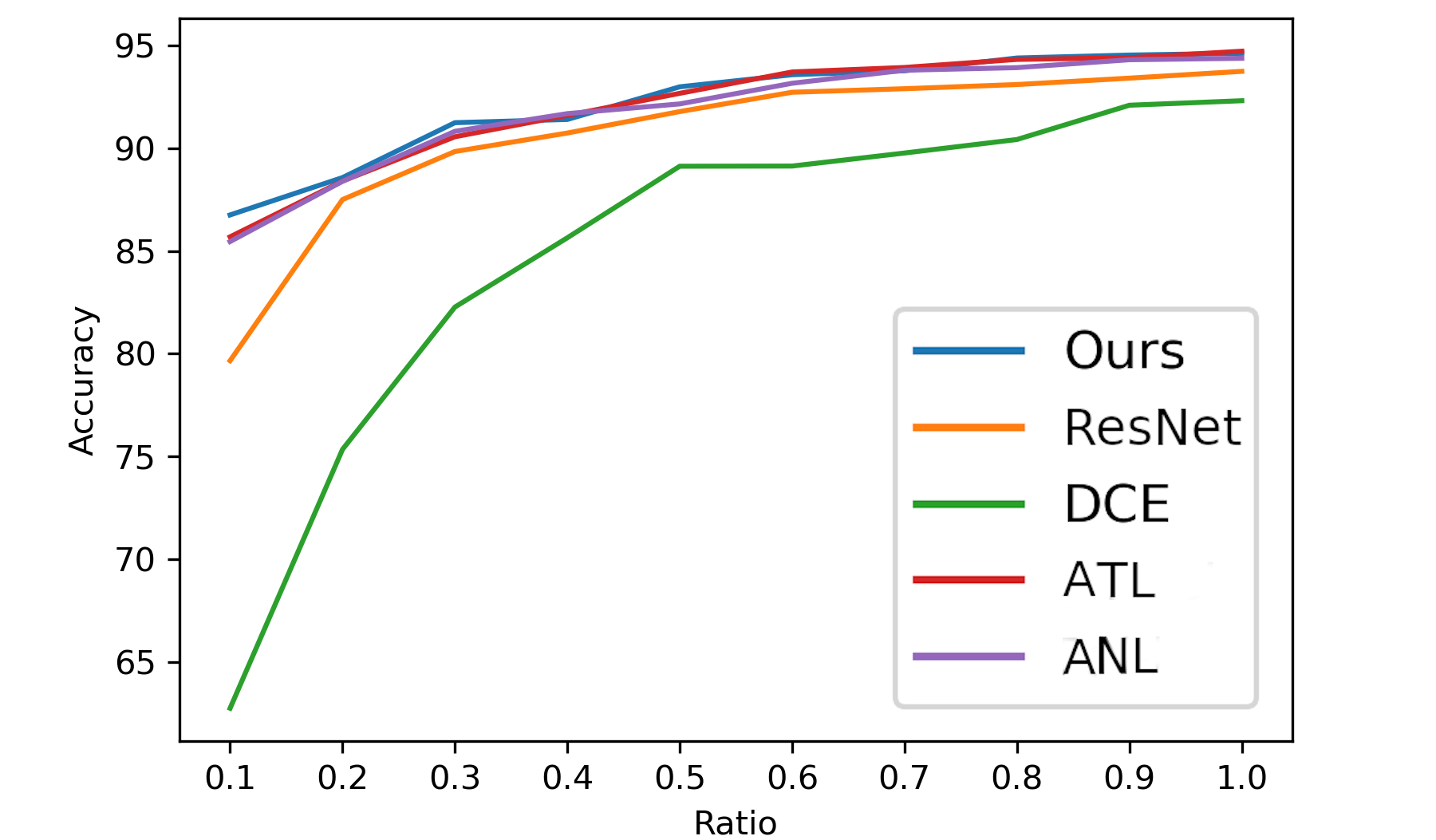}\caption{}\label{subfig:sample_efficiency}
\end{subfigure}
\begin{subfigure}{0.44\textwidth}
\begin{minipage}[t]{\textwidth}
\includegraphics[width=1\textwidth]{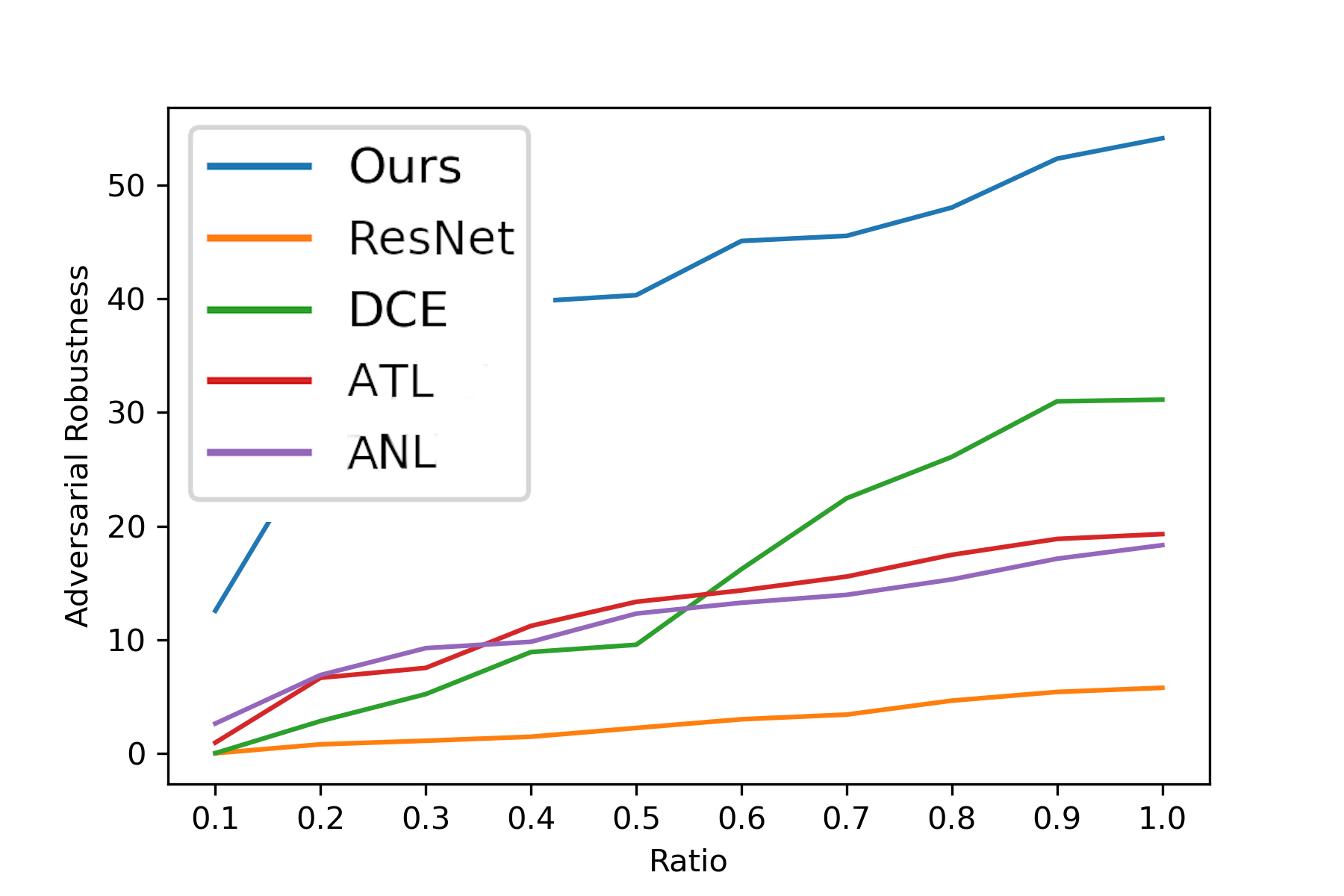}\caption{}\label{subfig:adv_efficiency}
\end{minipage}
\end{subfigure}
\centering
\caption{
Sample efficiency test on SVHN dataset. \cref{subfig:sample_efficiency} has the Accuracy and \cref{subfig:adv_efficiency} the Adversarial Robustness on FGSM attack with respect to different ratios of training data. Compared to other prototypical frameworks (DCE) our approach has significantly improved accuracy and is at par with other contrastive learning approaches. Our method out-perform all other approaches in `clean' adversarial robustness on FGSM for limited training data.}
\label{figure:sample_efficiency}
 \end{figure*}
We evaluate the ``Clean'' accuracy on the unperturbed test set and the accuracy on the same test set after it has been perturbed by FGSM, PGD, BIM, PGD with random start in  $l_{\infty}$ space ``$\text{PGD}_{RS}$'', PGD in $l_2$ space ``$\text{PGD}_{L2}$'' and AutoAttack. For all baselines we evaluate the robustness of multiple checkpoints and report the numbers from the best performing checkpoint. 
The results in \cref{table:adv_robustness} demonstrate that our method outperforms other baselines for all five types of attacks with a larger improvement on stronger attack settings (PGD, BIM, $\text{PGD}_{RS}$) and more complex datasets such as for CIFAR and SVHN. The improved robustness is larger for stronger attacks in $l_\infty$. Furthermore, when evaluated in AutoAttack, our model is the only that can achieve 0.1109 \textit{robustness}, as compared to 0.0 for all other baselines.
In agreement with previous work \cite{tsipras2018robustness} we observe a trade-off between clean accuracy and robustness. All model can reach a clean accuracy comparable to previously reported benchmarks \cite{he2016deep} but at the cost of robustness. When evaluating a model at the check-point of peak robustness the clean accuracy significantly suffers for non-robust model. Where the clean accuracy drops from 92.2\% to 82.76\% for a ResNet backbone. 
\\
\textbf{Out-of-Distribution} (OOD) 
We train all baselines and our own model on SVHN as $\mathcal{D}^\emph{in}$  and calculate the ability to detect out-of-distribution datasets such as ImageNet \cite{le2015tiny}, LSUN \cite{yu2015lsun} and Gaussian Noise using methodology from \cite{hendrycks2018deep} and metrics from \cite{liang2017enhancing}. We use a normal distribution for Gaussian Noise, normalize and clip the values to be within the same range as SVHN.  Results are presented in \cref{table:ood_svhn}. Based on Detection Error, our method surpasses the second-best approaches by 6.1\%  for ImageNet, 44\% for LSUN, and 4.8\% for Gaussian noise out-of-distribution datasets. We perform dimensionality reduction (t-SNE) \cite{tsne} on the feature representation and visualize the inter-class separability to Gaussian noise outliers in supplementary fig. 2 
and attach a larger size figure in our supplementary. From the figure it can be observed that most outliers are outside the classification boundary of the prototype. \\
\textbf{Attack Detection} To further demonstrate the robustness of our framework, we test our model and baselines with the same methodology as in the OOD detection experiment. We use PGD to generate adversarial samples as $\mathcal{D}^{out}$. We use CIFAR10 and SVHN as $\mathcal{D}^{in}$. The experiments evaluate the robustness to detect if a test sample has been perturbed by an adversarial attacker. Results are shown in \cref{table:ood_pgd}. Our method significantly outperforms all baselines for every evaluation metric. The detection error for all the baseline methods is 50\% with one exception. Since OOD detection is a binary classification task, the results indicate the baselines have no discriminative power in this setting. Our method is the only framework that can detect adversarial samples.
\begin{table}[h]
\begin{center}
\caption{Adversarial attack detection on CIFAR10 and SVHN datasets.}
\label{table:ood_pgd}
\scalebox{0.7}{
\begin{tabular}{@{}clccccc@{}}
\toprule
                                                                                & \multicolumn{1}{c}{} & \textbf{\begin{tabular}[c]{@{}c@{}}FPR\\ (95\% TPR)\\ $\downarrow$\end{tabular}} & \textbf{\begin{tabular}[c]{@{}c@{}}Detection\\ Error\\ $\downarrow$\end{tabular}} & \textbf{\begin{tabular}[c]{@{}c@{}}AUROC\\ \\ $\uparrow$\end{tabular}} & \textbf{\begin{tabular}[c]{@{}c@{}}AUPR\\ In\\ $\uparrow$\end{tabular}} & \textbf{\begin{tabular}[c]{@{}c@{}}AUPR\\ Out\\ $\uparrow$\end{tabular}} \\ \midrule
\multirow{5}{*}{\textbf{\begin{tabular}[c]{@{}c@{}}CIFAR\\ (PGD)\end{tabular}}} & ResNet              & 100.0\%                                                               & 50.0\%                                                                 & 27.9\%                                                        & 33.5\%                                                         & 5.1\%                                                           \\
                                                                                & DCE                  & 97.9\%                                                                & 50.0\%                                                                 & 22.4\%                                                        & 34.5\%                                                         & 32.3\%                                                          \\
                                                                                & ATL                  & 100.0\%                                                               & 50.0\%                                                                 & 40.4\%                                                        & 35.8\%                                                         & 0.5\%                                                           \\
                                                                                & ANL                  & 100.0\%                                                               & 50.0\%                                                                 & 26.7\%                                                        & 33.2\%                                                         & 2.8\%                                                           \\ \cmidrule(l){2-7}
                                                                                & \textbf{Ours}        & \textbf{89.9\%}                                                       & \textbf{37.3\%}                                                        & \textbf{66.5\%}                                               & \textbf{67.1\%}                                                & \textbf{62.8\%}                                                 \\ \midrule
\multirow{5}{*}{\textbf{\begin{tabular}[c]{@{}c@{}}SVHN\\ (PGD)\end{tabular}}}  & ResNet              & 100.0\%                                                               & 50.0\%                                                                 & 38.5\%                                                        & 35.3\%                                                         & 1.8\%                                                           \\
                                                                                & DCE                  & 89.1\%                                                                & 41.4\%                                                                 & 50.7\%                                                        & 45.6\%                                                         & 55.8\%                                                          \\
                                                                                & ATL                  & 99.9\%                                                                & 50.0\%                                                                 & 6.8\%                                                         & 31.5\%                                                         & 27.2\%                                                          \\
                                                                                & ANL                  & 99.7\%                                                                & 50.0\%                                                                 & 12.0\%                                                        & 32.4\%                                                         & 32.2\%                                                          \\ \cmidrule(l){2-7}
                                                                                & \textbf{Ours}        & \textbf{76.1\%}                                                       & \textbf{38.2\%}                                                        & \textbf{66.0\%}                                               & \textbf{68.8\%}                                                & \textbf{65.5\%}                                                 \\ \bottomrule
\end{tabular}
}
\end{center}
\end{table}

\subsubsection{Sample efficiency}

We evaluate the ability of each model to efficiently learn from training samples. We randomly sample a subset of the SVHN dataset and train multiple models as we vary the sample size. 
We evaluate and report the best accuracy score and adversarial robustness for an FGSM attack on each model on the entire test set. The increase in accuracy and robustness as the training set ratio increases can be found in  \cref{figure:sample_efficiency}. Our method is highly robust even with limited training samples. SCPL achieves a 0.87 test accuracy score using only 10 percent of the full training dataset in contrast to 0.6 for a ResNet-18 backbone. We calculate the Return on Investment (ROI) on the training dataset size using the dataset size as cost and adversarial robustness increase as the net gain. SCPL has an ROI of 164\% with an increase in robustness from 12.5\% to 28.9\% and an increase in training samples from 0.1 to 0.2 percent as opposed to the next best model for the same ratio interval, ANL, with an ROI of 33\%.
The robustness of SCPL increases faster compared to other baselines. 

\subsubsection{Ablation studies}\label{sec:ablation} We perform extensive ablation studies on the hyper-parameters of our method using two datasets, CIFAR10 and SVHN. Results are presented on table 1 and table 2 in the Appendix
for CIFAR and SVHN dataset respectively. We train our model with identical training settings for different regularization coefficients $\lambda$ and numbers of prototypes per class $C$, with the similarity metrics for train and inference denoted as ( Train / Inference).
We denote the base model as `base' which is identical to the Adversarial Robustness experimental set-up with the only difference being that we train for 100 epochs. The results suggest the benefit of $\mathcal{L}_{norm}$ is that the features are gathered closer to their intra-class prototypes. We experiment with a cosine similarity which is the Dot-Product (``DP'') loss function for both inference and training. In agreement with our theoretical results, we conclude that ``DP'' can improve the robustness. In addition, we found that increasing the number of prototypes per class will degrade the model accuracy, which agrees with previous work \cite{DCE}.
We hypothesize that too many prototypes can be a difficult optimization objective and the reduced robustness is due to different prototypes providing a larger $p$-Neighbourhood for $S_i$ as we discussed in \textit{Provably Robust Framework} \cref{sec:method}.

\section{Discussion}
\label{sec:discussion}
Our proposed method outperforms the previous state-of-the-art, but there are limitations and potential future improvements. First, the prototypes are initialized with random weights in the current setting, which can lead to poor convergence at the beginning of the training stage. Warm-up methods could further improve model convergence \cite{arthur2006k}.

In addition, the similarity function between feature representations and the decision boundary $\alpha$ from theorem \textit{Provably Robust Classification} in the Appendix can be improved. Related to this is the design of an OOD method specific to our model.

Finally, the inference of our proposed method is based on prototype matching by assigning test examples to the class of their closest prototype, which cannot be directly applied to tasks where a continuous value needs to be predicted (i.e regressional analysis).

\section{Conclusion}
We propose a supervised contrastive prototype learning (SCPL) framework for learning robust feature representation. We replace the categorical classification layer of a DNN with a Prototype Classification Head (PCH). PCH is an ensemble of prototypes assigned to each class. We train PCH jointly with the DNN using a single optimization objective, N-Prototype Loss. We empirically validate our results and provide a theoretical justification for the improved robustness. We achieve increased robustness without requiring auxiliary data such as adversarial or out-of-distribution examples. Lastly, SCPL requires no modification to the classification backbone and thus is compatible with any existing DNN.

\clearpage
{\small
\bibliography{bib_main}
}

\clearpage
\appendix

\begin{appendices}

\theoremstyle{definition}

\newtheorem{definition}{Definition}

\newtheorem{theorem}{Theorem}

\section{Theoretical Results}

\textbf{Provably Robust Framework} A categorical classifier learns a decision boundary on a feature space that has the same dimensionality as the size of the classification task. For example, a binary classification task would produce a two-dimensional feature space for which each dimension is a feature that represents the probability that a sample belongs to each respective class. We identify the cause of the robustness-accuracy trade-off \cite{tsipras2018robustness} to be the direct relationship between the feature space dimensionality restricted by the number of classes (``problem size'').

For our analysis we define a powerful transformation \(f(X)=H\) where $X=\{x_i | i \in N\}$ is a sample with i.i.d features $x_i$ such that $\bigcap X = \emptyset$  and \(H=\{h_i | i \in D\}\) is the hidden representation. It follows that for any hidden representation of lower dimensionality s.t. $D<N$ a transformation function will be unable to produce a one-to-one mapping between $X$ and $H$ and thus $h_i$ can not be i.i.d. such that $\bigcap H > \emptyset$.

Consider a downstream categorical classifier that assign a sample $X_j$ to the class $k=\text{argmax}(f(X_j))$. Assume a strong $l_\infty$-norm adversary that can perturb $X$ and is bounded by a constant $\varepsilon$. Perturbations to the input space will lead to perturbations in the hidden representation $f(X_j)=h$ where $h\in \mathbb{R}^K$ for a problem with $K$ classes. 
A non-robust classifier will assign a smaller value for the hidden representation $h_k$ and for the true class $k$ of $X_j$. A robust classifier will be invariant to such noise and $h_k$ will be significantly greater than all other elements of the $H$-set such that $h_k\gg {H}\backslash\{h_k\}$.

A categorical classifier has a hidden space dimensionality $D$ equivalent to $K$. For any general transformation, as the hidden space dimensionality approaches the input space dimensionality $D\rightarrow N$, an optimal transformation can learn \(\bigcap H\rightarrow \emptyset\) and the inter-class distance increases such that $h_k\gg {H}\backslash\{h\}$. 
We thus identify that the issue of inter-class separation is due to the restricted hidden representation space for a high dimensional input space.

PCH has a flexible $D$ independent of the problem size dimensionality. This allows us to show that the improved performance is a consequence of the flexible dimensionality of the learned feature representation for our method.
Theorem \ref{thm:robust_classification} can explain the improved robustness as the number of dimensions of the feature space increases. Additionally, we conclude that the similarity metric used is important in creating provably robust classifiers. Although the theorem does not extend to $l_1$ distance, additional conditions on the similarity boundary can be applied using the triangle inequality. In this work, we do not focus on the design of the similarity function or approximation of the decision boundary and consider it an open problem. 

A consequence of \cref{thm:robust_classification} is that it provides a guarantee of robustness only based on a known classification set $S_i$. A perturbation $\delta$ that is similar to a sample or a sample that is similar $\delta$ will also be in the $\alpha$-Neighborhood of $S_i$. Any such perturbation could cause our theoretical framework to become non-robust to $\delta$. 
In practice, this would be equivalent to a data-poisoning attack on classifier. Such classifier will be trained to assign class $i$ to random noise. Moreover, it would be possible to construct an adversarial example using a $\delta$ perturbation that belongs to any class $i$, since $sim(\hat{\delta},m_i)> \alpha$. In practice, $D$ is limited by the amount of computational resources but as $D$ increases, such perturbation will craft adversarial examples that are perceptually different than the true class or original image. Thus, the strength of our theoretical framework is that any successful adversarial sample would not be perceptually similar to the original input for a classifier trained on natural images and for really large $D$. 

Our experiments corroborate the theoretical implications. We can increase the robustness of the model by increasing $D$ and 
models robust to Gaussian noise are also robust to adversarial attacks. We provide detailed proofs in the remainder of this section and additional experiments in \cref{sec:addtl_exp}.

\begin{definition}[Classification Similarity Function] $sim_f$ compute a metric between two vectors  $x,m$ $\in R^D$, is distributive over vector addition $sim_f(x+\delta,m)=sim_f(x,m)+sim_f(\delta,m)$, is communicative  $ sim_f(x,m)=sim_f(m,x)$, and is divergent as the number of dimensions increases such that \( \lim_{D\to\infty} sim_f(x,m) = \pm \infty \)
\end{definition}
\begin{definition}[Classification Set] The set $\mathcal{S}_i$ is the set of samples $x\in R^D$ for prototype vector $\bm{m}_i$ where $ sim_f(x,m)>\alpha$ given a positive \textit{similarity boundary} $0<\alpha<\infty$ such that $\mathcal{S}_i=\{x | sim_f(x,\bm{m}_i)>\alpha\}$
\end{definition}
\begin{theorem}[Provably Robust Classification]
\label{thm:robust_classification}
Given $ sim_f $ there exists a similarity boundary $\alpha$ and a prototype vector $m_i \in R^D$ for which there is no perturbation $\delta \in R^D$ such that the perturbed feature vector $h_x+\delta$ is misclassified with $ sim_f(m,h_x+\delta) < \alpha$ but $h_x$ is correctly classified with $ sim_f(x,m)>\alpha$ as the number of feature space dimensions $D$ increases with $D\rightarrow\infty$.

\end{theorem}

Proof:
Let $sim(\cdotp)$ be a \textit{classification similarity function} and let $x, m \in R^D$ where $x$ is in the classification set $\mathcal{M}=\{x | sim(x,m)>\alpha\}$ with $\alpha$ the similarity boundary. Let $\delta$ be a perturbation s.t. $sim(m,\delta) < \alpha$.

Assume that there exists a $\delta$ s.t. $sim(x+\delta,m)<\alpha$ but $sim(x,m)>\alpha$  (1) 

By the distributive property we have 
$sim(x,m)+sim(\delta,m)<\alpha$, 
$sim(x,m)<\alpha-sim(\delta,m)$ 

Let $\rho=\alpha-sim(\delta,m)$.

There exist an $\alpha$ such that $\rho>\alpha$, by substitution $\alpha-sim(\delta,m)<\alpha$ and thus $0<sim(\delta,m)$

By definition $sim$ is divergent as $D\rightarrow\infty$ with $0<sim(\delta,m)=+\infty$ and $0<\alpha<sim(m,x)=+\infty$ but $sim(m,x)+sim(m,\delta)<\alpha$ which is a contradiction.

\section{Additional Experiments}
\label{sec:addtl_exp}
The issue of robustness can be viewed as an issue of generalization \cite{xu2012robustness}. We perform additional experiments to corroborate our theoretical findings. We use a VGG-11 \cite{he2016deep} model pre-trained on ImageNet as provided by the python library PyTorch \cite{NEURIPS2019_9015}. We compute the Gradient Activation map \cite{selvaraju2017grad} for all images in three datasets CIFAR-10 \cite{Krizhevsky09learningmultiple}, ImageNet-1M \cite{imagenet_cvpr09}, and Gaussian random noise. We compute the mean activation value of all images within each dataset. The \textit{mean activation maps} at \cref{figure:activation_maps} show artifacts of the training settings. He et al. \cite{he2016deep} resizes the images during on the shortest side to be 256 pixels during training. The images are then randomly cropped to (224x224). As a consequence, the corners of the original image (32 pixels) will appear less frequently during training. The model learns to ignore the corners as shown by the activation map and over-fit to the center of the image. When random noise is fed such that there are no salient features from which the classifier can correctly assign the class, the network makes predictions based on the image corners. We observe that the salient features the network uses to classify an image are the cropping dimensions. This is an artifact of the classifier overfitting to the augmentation technique used during training. 

We analyze the prediction scores for a categorical classification model to find that it makes overconfident predictions for the wrong class. Our method (PCH) assigns high confidence on the in-distribution ($\mathcal{D}^\emph{in}$) dataset CIFAR-10 as can be seen by the area of the red curve in \cref{fig: std_cifar10}. Out-of-distribution images are assigned less confident prediction scores for our method. 

In contrast, a ResNet-18 model pretrained on ImageNet assigns high confidence predictions to out-of-distribution data. Additionally, the majority of predictions are for a single class. The majority predicted class is the same for a dataset between different runs and random iterations but different between datasets, as can be seen in \cref{fig:freq_gaus}.

Lastly, our model (PCH) has high inter-class separation. Out-of-distribution samples are distanced from the class centroids, as can be seen in \cref{figure:tsne}.

\newpage
\clearpage
\begin{table*}[ht]
\caption{Ablation study on \textbf{CIFAR10} dataset with hyper-Parameters explained in section 5.2.3, `Ablation Studies', 
of the main text.}
\label{table:abl_cifar}
\begin{center}
\begin{tabular}{@{}lcccccc@{}}
\toprule
\textbf{}    & \textbf{Clean} & \textbf{FGSM} & \textbf{PGD} & \textbf{BIM} & $\textbf{PGD}_{RS}$ & $\textbf{PGD}_{L2}$ \\ \midrule
base         & 88.980         & 27.370        & 5.240        & 7.340        & 5.870          & 68.090         \\ \midrule
$\lambda$=0        & 88.630         & 27.500        & 4.120        & 6.230        & 4.170          & 70.720            \\
$\lambda$=0.2          & 88.770         & 28.650        & 6.740        & 10.070       & 6.700          & 73.250      \\ \midrule
D=2          & 62.420        & 14.200        & 1.020        & 1.030        & 3.000         & 39.360         \\
D=8          & 83.080        & 19.540        & 1.540        & 1.700        & 3.730          & 55.680         \\\midrule
C=2          & 88.120         & 29.140        & 6.800        & 9.360        & 6.470          & 67.990         \\
C=5          & 85.810         & 23.660        & 5.040        & 6.870        & 4.560          & 66.310         \\
C=10         & 83.800         & 19.950        & 3.780        & 5.370        & 3.770          & 62.850         \\ \midrule
$l_1$  / $l_2$  & 83.600         & 23.490        & 5.280        & 7.220        & 5.480          & 64.930         \\
DP/ $l_2$  & 84.100         & 22.730        & 4.500        & 5.030        & 5.320          & 68.450         \\ \midrule
DP / DP & 82.260         & 19.470        & 3.500        & 4.950        & 3.580          & 63.950         \\
$l_1$ / $l_1$ & 86.340         & 23.940        & 5.600        & 5.760        & 6.230          & 70.060         \\ \midrule
$l_2$ / $l_1$      & 88.980         & 27.410        & 5.260        & 7.330        & 5.620          & 68.090         \\
$l_2$ / DP      & 88.980         & 27.410        & 5.260        & 7.350        & 5.810          & 68.080         \\ \bottomrule
\end{tabular}
\end{center}
\end{table*}

\begin{table*}[ht]
\caption{Ablation study on \textbf{SVHN} dataset with hyper-Parameters explained iin section 5.2.3, `Ablation Studies', 
of the main text.}
\label{table:abl_svhn}
\begin{center}
\begin{tabular}{@{}lcccccc@{}}
\toprule
\textbf{}    & \textbf{Clean} & \textbf{FGSM} & \textbf{PGD} & \textbf{BIM} & $\textbf{PGD}_{RS}$ & $\textbf{PGD}_{L2}$ \\ \midrule
base         & 94.872         & 52.524        & 8.547        & 17.048       & 10.257         & 85.802         \\ \midrule
$\lambda$=0        & 94.683         &   50.980     & 6.550        & 15.946       & 9.135          & 86.086          \\
$\lambda$=0.2          & 94.780         & 53.373         & 10.483       & 19.119       & 12.412         &  86.616       \\\midrule

D=2          & 79.122         & 42.102       & 6.822        & 8.332       & 8.397          & 62.780         \\
D=8          & 92.617         & 23.660        & 5.040        & 6.870        & 4.560          & 66.310         \\\midrule
C=2          & 94.610         & 54.099        & 15.112       & 24.804       & 17.083         & 86.209         \\
C=5          & 94.434         & 48.986        & 8.805        & 17.705       & 10.910         & 85.510         \\
C=10         & 94.073         & 45.164        & 6.457        & 13.299       & 8.313          & 84.342         \\ \midrule
$l_1$ / $l_2$  & 94.415         & 47.945        & 8.451        & 17.037       & 10.026         & 84.930         \\
DP / $l_2$  & 95.022         & 51.314        & 6.200        & 18.758       & 8.397          & 87.519         \\ \midrule
DP / DP & 93.654         & 53.619        & 8.831        & 17.920       & 10.191         & 83.171         \\
$l_1$ / $l_1$ & 94.626         & 53.357        & 6.523        & 18.946       & 9.043          & 87.062         \\ \midrule
$l_2$ / $l_1$      & 94.872         & 52.520        & 8.582        & 17.098       & 10.360         & 85.802         \\
$l_2$ / DP      & 94.872         & 52.520        & 8.586        & 17.098       & 10.345         & 85.802         \\ \bottomrule
\end{tabular}
\end{center}
\end{table*}
\newpage 
\clearpage
\begin{figure*}[ht]
\centering
\begin{subfigure}[t]{0.3\textwidth}
\includegraphics[width=1\linewidth]{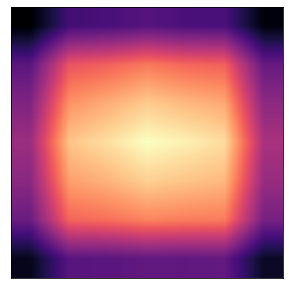}
\caption{ImageNet Activation Map}
\label{fig:imgnets_grad}
\end{subfigure}
\begin{subfigure}[t]{0.3\textwidth}
\includegraphics[width=1\linewidth]{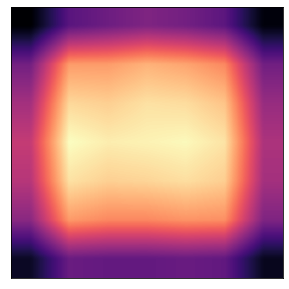}
\caption{CIFAR-10 Activation Map}
\label{fig:cifar10_grad}
\end{subfigure}
\begin{subfigure}[t]{0.3\textwidth}
\includegraphics[width=1\textwidth]{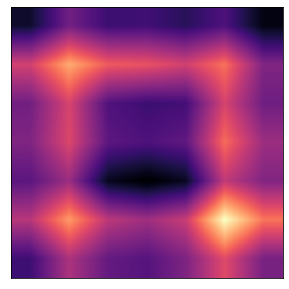}
\caption{Random Noise Activation Map}
\label{fig:rand_grad}
\end{subfigure}
\centering
\caption{We obtain the Gradient Activation map from a VGG11 categorical classification model that is pre-trained on ImageNet. We generate the map for three datasets, ImageNet \cref{fig:imgnets_grad}, CIFAR-10 \cref{fig:cifar10_grad}, Gaussian Noise \cref{fig:rand_grad}. The activation map for each image in the dataset is computed and then aggregated by the mean value for each pixel on the original image. Higher values for a pixel correspond to higher importance when predicting the corresponding class.} 
\label{figure:activation_maps}
\end{figure*}

\begin{figure*}[ht]
\centering
\includegraphics[width=\textwidth]{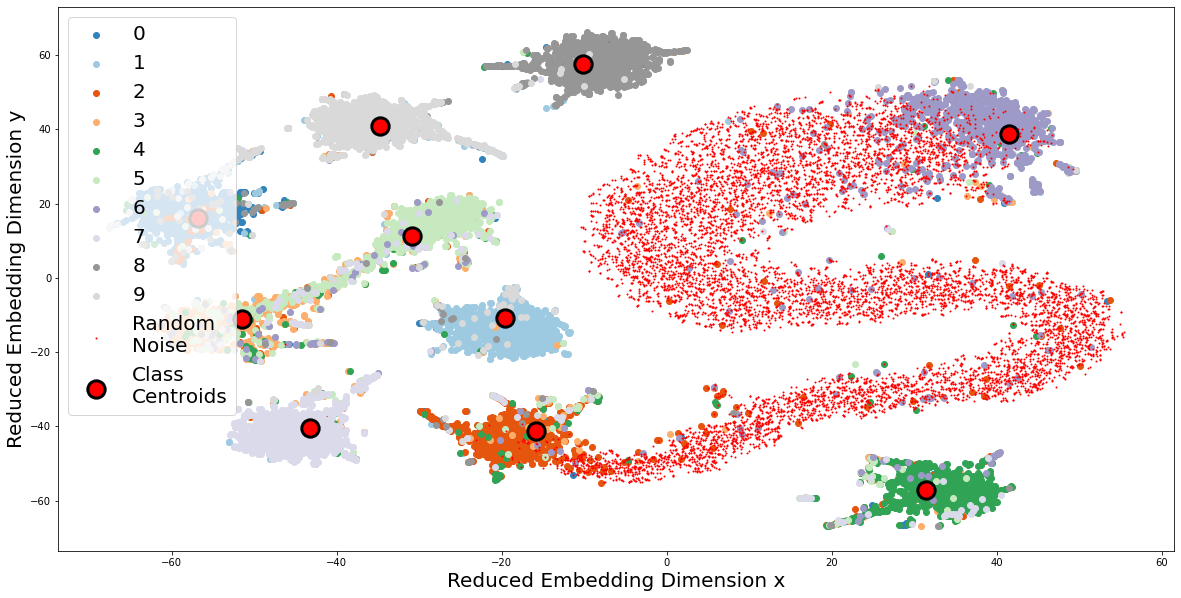}

\centering
\caption{Dimensionality reduction (t-SNE) on the embedding space learned by our method (PCH) on the CIFAR-10 dataset. The smaller red markers represent the random noise. The big red markers are the class centroids. Each of the CIFAR10 classes is assigned a different color.} 
\label{figure:tsne}
\end{figure*}

\newpage 
\clearpage
\begin{figure*}[ht]
\centering
\begin{subfigure}[t]{.5\linewidth}
\includegraphics[width=1\textwidth]{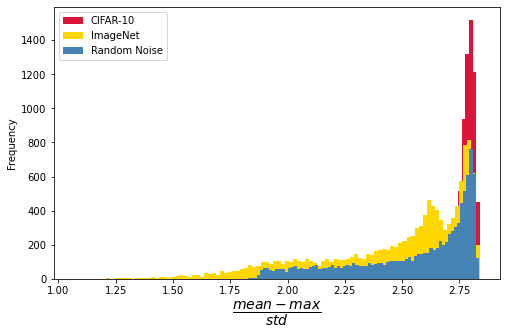}
\caption{PCH model trained on CIFAR-10 $D_{in}$ with Gaussian Noise and ImageNet as $D_{out}$}
\label{fig: std_cifar10}
\end{subfigure}

\begin{subfigure}[t]{.5\linewidth}
\includegraphics[width=1\textwidth]{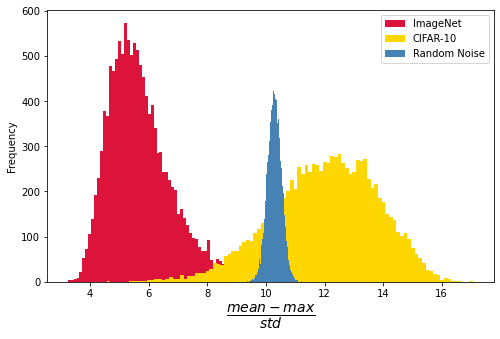}
\caption{Categorical model trained on ImageNet ($D_{in}$)  with Gaussian Noise and CIFAR-10 as $D_{out}$}
\label{fig: std_imgnet}
\end{subfigure}
\begin{subfigure}[t]{.5\linewidth}
\includegraphics[width=1\textwidth]{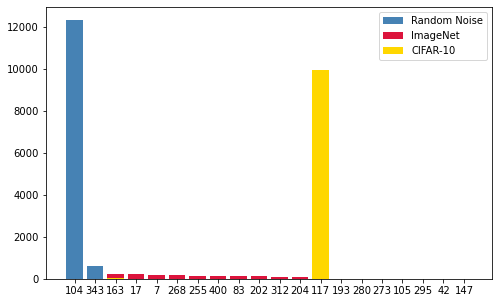}
\caption{Predicted class frequency counts for categorical model trained on ImageNet ($D_{in}$) with Gaussian Noise and CIFAR-10 as $D_{out}$}
\label{fig:freq_gaus}
\end{subfigure}
\caption{We examine the prediction confidence score for out of distribution dataset. The prediction scores for CIFAR-10 \cref{fig: std_cifar10}, ImageNet \cref{fig: std_imgnet} and Random Noise \cref{fig:freq_gaus} are computed. For the prediction vector from each sample we compute the mean, maximum, and standard deviation to calculate $\rho=\frac{max-mean}{std}$. Higher $\rho$ values for $\mathcal{D}^\emph{in}$ relative to $\mathcal{D}^\emph{out}$ are better. The $\rho$ curve mass density for PCH is higher when compared to both random noise and ImageNet. The same does not apply for \cref{fig: std_imgnet} where the classifier is more confident in predicting random noise than it is in predicting $\mathcal{D}^\emph{in}=\text{ImageNet}$} 
\label{figure:pred_scores}
\end{figure*}

\clearpage

\end{appendices}

\end{document}